\documentclass{article}

\PassOptionsToPackage{numbers}{natbib}
\usepackage[final]{nips_2017}

\usepackage[utf8]{inputenc} % allow utf-8 input
\usepackage[T1]{fontenc}    % use 8-bit T1 fonts
\usepackage{hyperref}       % hyperlinks
\usepackage{url}            % simple URL typesetting
\usepackage{booktabs}       % professional-quality tables
\usepackage{amsfonts}       % blackboard math symbols
\usepackage{amsmath}
\usepackage{nicefrac}       % compact symbols for 1/2, etc.
\usepackage{microtype}      % microtypography
\usepackage{bm}
\usepackage{tabularx}
\usepackage{graphicx}
\usepackage{wrapfig}
\usepackage{caption}
\usepackage{subcaption}
\usepackage{float}

\newcommand{\vect}[1]{\bm{#1}}
\newcommand{\matr}[1]{\bm{#1}}

\newcommand{\vc}[0]{\vect{c}}

\newcommand{\vh}[0]{\vect{h}}

\newcommand{\vx}[0]{\vect{x}}
\newcommand{\vz}[0]{\vect{z}}

\newcommand{\vs}[0]{\vect{s}}

\newcommand{\vl}[0]{\vect{l}}

\newcommand{\vo}[0]{\vect{o}}
\newcommand{\vy}[0]{\vect{y}}

\newcommand{\vr}[0]{\vect{r}}

\newcommand{\mW}[0]{\matr{W}}

\newcommand{\mE}[0]{\matr{E}}

\newcommand{\mU}[0]{\matr{U}}

\newcommand{\mV}[0]{\matr{V}}

\title{Modulating and attending the source image during encoding improves Multimodal Translation}

\author{
  Jean-Benoit Delbrouck\\
  University of Mons, Belgium\\
  \texttt{jean-benoit.delbrouck@umons.ac.be} \\
  \And
   St\'ephane Dupont\\
  University of Mons, Belgium\\
  \texttt{stephane.dupont@umons.ac.be} \\
}

\begin{document}

\maketitle

\begin{abstract}
  We propose a new and fully end-to-end approach for multimodal translation where the source text encoder modulates the entire visual input processing using conditional batch normalization, in order to compute the most informative image features for our task.
  %Not only we report a first attempt on the multimodal translation task where the convolutional mechanism is part of the trainable network, but
  Additionally, we propose a new attention mechanism derived from this original idea, where the attention model for the visual input is conditioned on the source text encoder representations. In the paper, we detail our models as well as the image analysis pipeline. Finally, we report experimental results. They are, as far as we know, the new state of the art on three different test sets.
\end{abstract}

\vspace{-4mm}
\section{Introduction} \label{intro}

Since the first multimodal machine translation (MMT) shared task \citep{specia-EtAl:2016:WMT} has been released, the community struggled to prove the effectiveness of images in the translation process. Most of the works \citep{caglayan-EtAl:2016:WMT,huang-EtAl:2016:WMT,calixto-elliott-frank:2016:WMT,delbrouck-dupont:2017:EMNLP2017} naturally focused on using a soft attention mechanism \citep{BahdanauCB14} on the convolutional features (also called attention maps) of an image, alongside with a textual attention mechanism, because this approach has shown great success in image captioning \citep{icml2015_xuc15}. First attempts were relatively unsuccessful (i.e. slightly lower than a strong monomodal baseline) and it was hard to figure out the real reasons of these underwhelming results. The last multimodal translation shared task \citep{ElliottFrankBarraultBougaresSpecia2017} decided to address this issue by releasing two new test-sets containing pictures of new Flickr groups and sentences with ambiguous verbs so we know for sure the image could play a disambiguation role in the translation process. At the same time tough, the monomodal baseline got stronger and stronger with new findings regarding recurrent network architectures, such as layer normalization \citep{ba2016layer}, making the improvements brought by a new modality thiner. The most successful recent try \citep{caglayan-EtAl:2017:WMT} focused on using the max-pooled features extracted from a CNN to modulate some components of the system (i.e. the target embeddings). So far, researchers extract the image features from a pre-trained CNN without any intervention from the encoder-decoder model used for translation.

We decide to take the leap and to modulate the feature extraction process by the linguistic input. More precisely, this paper aims to :

\begin{itemize}
\item Give a first try on a fully end-to end (visual and textual) multimodal translation model;
\item Condition the forward pass of the CNN to extract visual features according to the textual encoder;
\item Propose an encoder-based image attention model as opposed to the conventional attention mechanism used during decoding time;
\end{itemize}
%Research areas at the very intersection of computer vision and natural language processing is starting to draw more interest. Visual Question Answering (VQA) is one of those  \citep{fukui-EtAl:2016:EMNLP2016,Goyal_2017_CVPR}.%
In the area of NMT, two works are related to ours, in the sense that one modality analysis process is affected by the analysis of the other modality. Firstly, \citep{DBLP:journals/corr/ElliottK17} proposed an architecture with an encoder shared between two decoders : one to output a translated sentence and one to reconstruct (imagine as the authors say) the image features. The encoder was thus trained to learn grounded representation. Secondly, \citep{delbrouck2017multimodal} used a grounded attention mechanism (referred as "pre-attention") where the image features were refined according to the encoder's representation of the source sentence.

\section{Monomodal (Text-based) MT model} \label{enc_dec}

Our model is based on an encoder-decoder architecture with attention mechanism \citep{BahdanauCB14}. The encoder is a bi-directional RNN with Gated Recurrent Unit (GRU) layers \cite{ChungGCB14,cho-al-emnlp14}. A forward RNN $\overrightarrow{f}_\text{enc}$ and a backward RNN $\overleftarrow{f}_\text{enc}$ both read an input sequence $x = (x_1, x_2, \hdots , x_M)$, ordered from $x_1$ to $x_M$ and from $x_M$ to $x_1$ respectively. Each RNN produces a hidden state $\widehat{\vh}_{i}$ for each word $x_i$. We create a sequence of annotations $h = (\vh_1, \vh_2, \hdots, \vh_M)$, $\vh_i = [\overrightarrow{\widehat{\vh}}_i;\overleftarrow{\widehat{\vh}}_i]$ where $[ \quad ]$ denotes the concatenation operation. Therefore, each annotation $\vh_i$ now contains the summaries of both the preceding words and the following words.

The decoder ${f}_\text{dec}$ is a CGRU (two stacked GRUs) that predicts the probability of a target sequence $y = (y_1, y_2, \hdots , y_K)$ based on $h$. At each decoding step $t$, an unnormalized
attention score $\widehat{a}_{i}$ is computed for each source annotation $\vh_i$ using the first GRU’s hidden state $\vs_t$ and $\vh_i$ itself (equation \ref{eqn:6}):
\def\tabularxcolumn#1{m{#1}}
\noindent\begin{tabularx}{\textwidth}{
    >{\hsize=1.4\hsize}X
    >{\hsize=0.8\hsize}X
    >{\hsize=0.8\hsize}X
  }
  \begin{equation}
\widehat{a}_{i} = \mV_a^T \text{tanh}( \mW_h \vh_i + \mW_s \vs_t)
    \label{eqn:6}
  \end{equation} &
  \begin{equation}
a_{i} = \frac{\widehat{a}_{i}}{\sum\nolimits_{m}^M\widehat{a}_{m}}
    \label{eqn:7}
  \end{equation} &
  \begin{equation}
\vc_t = \sum\limits_{i}^M a_{i} \vh_i
    \label{eqn:8}
  \end{equation}
\end{tabularx}
The attention vector $\vc_t $ is calculated as a weighted average of the source states $h$ as shown in equation \ref{eqn:8}. The second GRU computes the final state $\widehat{\vs}$ of the decoder with $\vc_t$ and $\vs_t$. The decoder outputs a distribution over a vocabulary of fixed-size V based on the recurrent state of the second GRU $\widehat{\vs}_t$, the previous words $y_{<t}$, and the attention vector $\vc_t$:
\noindent\begin{tabularx}{\textwidth}{@{}XX@{}}
  \begin{equation}
\vo_t = \text{tanh}(\vy_{t-1} + \mW_{\widehat{s}} \widehat{\vs}_t + \mW_c \vc_t)
    \label{eqn:9}
  \end{equation} &
  \begin{equation}
P(y_t | y_1, \hdots, y_{t-1}, x) = \text{softmax}(\mW_o \vo_t)
    \label{eqn:10}
  \end{equation} 
\end{tabularx}
The whole model is trained end-to-end by minimizing the negative log likelihood of the target words using stochastic gradient descent.

\section{Our multimodal MT model}

As stated in the introduction, the convolutional network extracting the image features is now part of the training procedure. We chose a residual network (ResNet) who iteratively refines a representation
by adding pass-through routing so that layers receive more detailed information rather than the information processed by the previous layer or adjacent to it. This modification enables to train deep convolutional networks without suffering too much from the vanishing gradient problem. 

\subsection{Residual Network} \label{resnet}
ResNets are built from residual blocks:

$$\vy = \mathcal{F}(\vx, \{\mW_i\}) + \mW_s\vx $$

Here, $\vx$ and $\vy$ are the input and output vectors of the layers considered. The function $\mathcal{F}(\vx, \{\mW_i\})$ is the residual mapping to be learned. For an example, if we consider two layers, $\mathcal{F} = \mW_2 \sigma ( \mW_1 \vx)$ where $\sigma$ denotes ReLu function. The operation $\mathcal{F} + \vx$ is the shortcut connection and consists of an element-wise addition. Therefore, the dimensions of $\vx$ and $\mathcal{F}$ must be equal. When this is not the case (e.g., when changing the input/output
channels), the $\mW_s$ matrix performs a linear projection by the shortcut connections to match the dimension. Finally, it performs a last second nonlinearity after the addition (i.e., $\sigma(y))$. A group of blocks are stacked to form a stage of computation. The general ResNet architecture starts with a single convolutional layer followed by 4 stages of computation.

\subsection{Conditional Batch Normalization } \label{cbn}

A ResNet adopts batch normalization (BN)\citep{ICML-2015-IoffeS} right after each convolution and before activation. This techniques tackles the problem of internal covariate shift (the distribution of each layer's inputs changes during training, as the parameters of the previous layers change) and addresses it by normalizing layer inputs :
\noindent\begin{tabularx}{\textwidth}{@{}XX@{}}
  \begin{equation}
  \widehat{\vx}^{(k)} = \frac{\vx^{(k)} - \text{E}_B[\vx^{(k)}]}{\sqrt{\text{Var}_B[\vx^{(k)}]} + \epsilon} 
    \label{eqn:1}
  \end{equation} &
  \begin{equation}
\vy^{(k)}=\gamma^{(k)}\widehat{\vx}^{(k)} + \beta^{(k)}
    \label{eqn:2}
  \end{equation} 
\end{tabularx}

The network applies the above equation \ref{eqn:1} to make each feature dimension $k$ of the input $\vx$ in the whole mini-batch follow a zero mean and unit variance Gaussian. On top of that, the model has the opportunity to shift and scale the result as shown in equation \ref{eqn:2} before going through the the non-linearity (ReLu). At inference time, the batch mean and variance are replaced by a single empirical mean and variance of activations during training. 

To modulate the visual processing by language, we will predict a small change in the shift and scale parameters of equation \ref{eqn:2} according to the text-based source annotations sequence $h$ as already been proposed in the related VQA task \citep{modul} (called "\textit{Modulated ResNet}" in the author's paper). We call this conditional batch normalization. To do so, we use a one-hidden-layer MLP to predict these deltas for all feature
maps within the layer:
\noindent\begin{tabularx}{\textwidth}{@{}XXX@{}}
  \begin{equation}
  \Delta \gamma, \Delta \beta = \text{MLP}(q(h))
    \label{eqn:3}
  \end{equation} &
  \begin{equation}
\widehat{\gamma}^{(k)} = \gamma^{(k)} + \Delta \gamma^{(k)}
    \label{eqn:4}
  \end{equation} &
  \begin{equation}
\widehat{\beta}^{(k)} = \beta^{(k)} + \Delta \beta^{(k)}
    \label{eqn:5}
  \end{equation}
\end{tabularx}

\vspace{-5mm}

where $q(\{\vh_1, \hdots, \vh_M \}) = \tanh \bigg( \mW_q \cdot \frac{1}{M} \sum\nolimits_{i=1}^M \vh_i \bigg)$
\vspace{-4mm}
\subsection{Image Features} \label{im_att_mech}
This section aims to explain which image features are extracted from the ResNet and how the model described in section \ref{enc_dec} uses it. Commonly, two types of features are useful for machine translation: global pooled features (a vector of features) and convolutional features (also called an attention map, a 3D-matrice). Because we use one or the other, our model now has two variants referred as "\textit{pool5}" and "\textit{conv}" in the results section.

\subsubsection{Global pool5 Features}

In the ResNet architecture, at the end of the 4th stage sits a max-pooling layer just before the fully connected layer whose output is a global 2048-dimensional visual representation $V$ of the image. We use $V$ to modulates each source annotation $\vh_i$ using element-wise multiplication (as done in \citep{caglayan-EtAl:2017:WMT}):
  \begin{equation}
\vh_i = \vh_i \odot \tanh (\mW_{\text{pool}} \cdot V)
\label{global}
  \end{equation}
Because $V$ is a vector of features, a "\textit{pool5}" model does not need a second attention mechanism.

\subsubsection{Convolutional Features} \label{conv_feat}

At the end of the ResNet 3rd stage, after the ReLu activation (res4f), we extract convolutional feature maps of 7x7x1024 (the 3D matrice) that are regarded as 49 spatial annotations of 1024-dimension each. We use a soft attention mechanism over the 49 visual spatial locations $(\vl_1, \hdots, \vl_{49})$ at each decoding step $t$. It is the exact same mechanism of section \ref{enc_dec} but with $\vh_i$ replaced by $\vl_i$:
\def\tabularxcolumn#1{m{#1}}
\noindent\begin{tabularx}{\textwidth}{
    >{\hsize=1.4\hsize}X
    >{\hsize=0.8\hsize}X
    >{\hsize=0.8\hsize}X
  }
  \begin{equation}
\widehat{a}_{i} = \mV_a^T \text{tanh}( \mW_l \vl_i + \mW_s \vs_t)
    \label{eqn:48}
  \end{equation} &
  \begin{equation}
a_{i} = \frac{\widehat{a}_{i}}{\sum\nolimits_{m}^{49}\widehat{a}_{m}}
    \label{eqn:49}
  \end{equation} &
  \begin{equation}
V_t = \sum\limits_{i}^{49} a_{i} \vl_i
    \label{eqn:50}
  \end{equation}
\end{tabularx}
  This hence constitutes an additional attention mechanism to the one described in section \ref{enc_dec}, applied to the visual annotations $\vl_i$ rather than the text annotations $\vh_i$. The weighted sum of the attention process of equation \ref{eqn:50} leaves us with a 1024-dimensional visual representation $V$ of the image. We then use it to modulates each source annotation $\vh_i$ as described in equation \ref{global}.

\section{Encoder-based image attention}\label{enc-based}

In machine translation, any image attention mechanism on convolutional features -- soft \citep{BahdanauCB14}, local or stochastic \citep{delbrouck-dupont:2017:EMNLP2017} -- happens on the decoder-side (based on $\vs_t$ as seen in the previous section \ref{conv_feat}). At each time-step $t$, the decoder has to decide which spatial features are interesting to decode the next translated token. However, when it comes to translate a sentence in real life, we rather tend to imagine a visual representation as soon as we read the source sentence. The encoder should probably be the strongest place to build a strong visual representations for our translation task. This hypothesis is reinforced by the additional role the encoder now endorse: modulating the visual processing as described in section \ref{cbn}. Because the encoder now plays a part in the making of these convolutional features, we propose to apply the attention mechanism for the visual representation during the encoding.

\begin{wrapfigure}{r}{0.5\textwidth}
\vspace{-6mm}
  \begin{center}
    \includegraphics[width=0.50\textwidth]{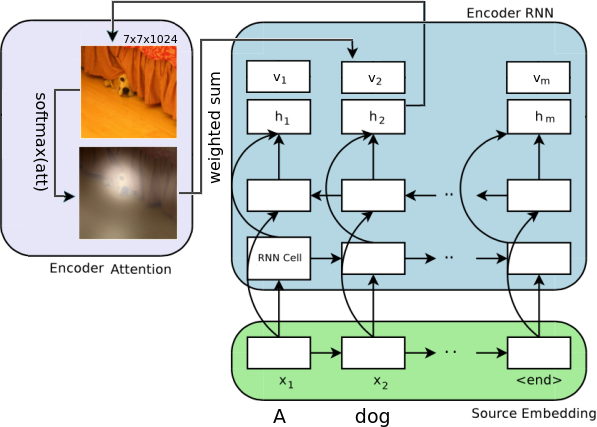}
  \end{center}
  \caption{Encoder-based attention for time-step 2}
  \label{enc_att}
\end{wrapfigure}

As shown in Figure \ref{enc_att}, the encoder now builds, at every time-step $i$, a textual representation $\vh_i$ and a visual representation $V_i$ of the word $x_i$. The encoder visual attention module is similar to the one described in subsection \ref{conv_feat}, but takes place in the encoder. Therefore, equation \ref{eqn:48} does not depend on the decoder state $\vs_t$ anymore but on the source annotation $\vh_i$. Now the decoder, at every decoding-step $t$, still computes a soft alignment over the source sentence annotations $\vh$ and hence gets, on its way, the visual representations $V$ as well. In contrast to earlier works on multimodal MT, the decoder is now equipped with only one multimodal attention mechanism, as the textual and visual representation of a word are computed in the encoder.

\vspace{-2mm}
\section{Dataset}
\vspace{-1mm}
We used the Multi30K dataset \cite{elliott-EtAl:2016:VL16} which is an extended version of the Flickr30K Entities. For each image, one of the English descriptions was selected and manually translated into German by a professional translator. As training and development data, 29,000 and 1,014 triples are used respectively. We dispose of three test sets to score our models. The flickr Test2016 and the Flickr Test2017 set contain 1000 image-caption pairs and the ambiguous MSCOCO test set \citep{ElliottFrankBarraultBougaresSpecia2017} 461 pairs.

\vspace{-2mm}
\section{Experiments and Results}
\vspace{-1mm}
Previous work \citep{delbrouck2017multimodal} showed that using visual features from different CNNs lead to variable translation performance for a same encoder-decoder model. In the this paper, we stick to two different versions of ResNet : the ResNet v1 detailed in our section \ref{cbn} and ResNet v2, a slight variant of ResNet v1 as described in \citep{He2016}. The image preprocessing operation is described in Appendix \ref{appendixA}. 

\begin{table*}[!ht]
		\centering
        		\caption{RN = ResNet, CBN = Conditional Batch Normalization, FT = fine tuning of the last ResNet stage and (*) is used as baseline }
        \hspace*{-0.9cm}
        \def\arraystretch{1.3}%  1 is the default, change whatever you need
		\begin{tabular}{lcccccc}
        \multicolumn{1}{c}{\bf Model}  &\multicolumn{2}{c}{\bf Test 2016}  &\multicolumn{2}{c}{\bf Test 2017}  &\multicolumn{2}{c}{\bf Ambiguous COCO}
			\\ \hline \\
			&BLEU$\uparrow$&METEOR$\uparrow$ &BLEU$\uparrow$&METEOR$\uparrow$ &BLEU$\uparrow$&METEOR$\uparrow$  \\ 
			 
			Pre-trained Pool5* \citep{caglayan-EtAl:2017:WMT}    
            &38.4 $\pm$ 0.3& 57.8 $\pm$ 0.5 & 31.1 $\pm$ 0.7 & 51.9 $\pm$ 0.2 & 27.0 $\pm$ 0.7 & 47.1 $\pm$ 0.7  \\
            
			RN v1 CBN Conv
           &38.9 $\pm$ 0.3& 57.1 $\pm$ 0.6 & 30.0 $\pm$ 1.1 & 50.9 $\pm$ 0.2 & 26.3 $\pm$ 0.9 & 46.5 $\pm$ 0.6  \\           
            
            RN v1 CBN Pool5
            &39.4$ \pm$ 0.8& \underline{\textbf{57.9}} $\pm$ 0.6 & \underline{\textbf{31.5}} $\pm$ 0.4 & 52.2 $\pm$ 0.5 & \underline{\textbf{27.4}} $\pm$ 0.9 & 48.1 $\pm$ 0.6  \\
           
           RN v2 CBN Pool5
            &38.7 $\pm$ 0.3& 56.5 $\pm$ 0.5 & 30.1 $\pm$ 0.7 & 51.1 $\pm$ 0.6 & 26.5 $\pm$ 0.7 & 46.3 $\pm$ 0.6  \\           
            
            RN v1 CBN FT Pool5
            &38.2 $\pm$ 0.6& 57.5 $\pm$ 0.6 & 30.4 $\pm$ 0.6 & 51.4 $\pm$ 0.7 & 26.4 $\pm$ 0.9 & 46.8 $\pm$ 0.4  \\            
            
            RN v1 CBN enc-att
            &\underline{\textbf{40.5}} $\pm$ 0.8& \underline{\textbf{57.9}} $\pm$ 0.6 & 31.4 $\pm$ 0.4 & \underline{\textbf{52.5}} $\pm$ 0.7 & 27.3 $\pm$ 0.9 & \underline{\textbf{48.5}} $\pm$ 0.4  
		\end{tabular}      
        \label{score-tabular}
	\end{table*}

Both ResNets are pretrained on ImageNet. Unless stated otherwise, ResNet parameters are frozen during training, including scalars $\gamma^{(k)}$ and $\beta^{(k)}$ from section \ref{cbn}. We use the metrics BLEU \cite{Papineni:2002} and METEOR \cite{meteor-wmt:2014} to evaluate the quality of our models' translations. We stop a training if there is no METEOR improvements on the dev-set for 10k steps.

First and foremost, we can notice that conditional batch normalization enhances our model translations, specifically when using the global features (\textit{RN v1 CBN Pool5}). Applying CBN at every ResNet stage lead to the best improvement (cfr. table \ref{score-tabular2} in Appendix C) but we also find that fine-tuning the last layer does not improve this performance (\textit{RN v1 CBN FT Pool5}). This result reinforce our main postulate that modulating the visual process by language enhance the quality of the translations.  

Secondly, when using decoder-based attention on convolutional features as described in section \ref{conv_feat}, the model performs poorly (\textit{RN v1 CBN Conv}). As stated in the introduction, it's not sure we have enough data to successfully train an attention model. Nevertheless, using the encoder-based attention (section \ref{enc-based}) palliates this gap. Indeed, both models \textit{ RN v1 CBN enc-att} and \textit{RN v1 CBN Conv} have very close results.

%These two reasons strengthen the hypothesis that conditioning and attending the image by the encoder improves Multimodal NMT.
Lastly, using a ResNet v2 slightly deteriorates the results. The key difference between the two architectures is the use of batch normalization before every weight layer. A more in-depth study of the model parameters and architecture might be needed to figure out the cause of this small drop. Another possible future work would be the use larger images as ResNet inputs (448x448) to enjoy convolutional features of 196 spatial locations, as this has shown great success in VQA.

\section{Acknowledgements}
This work was partly supported by the Chist-Era project IGLU with contribution from the Belgian Fonds de la Recherche Scientique (FNRS), contract no. R.50.11.15.F, and by the FSO project VCYCLE with contribution from the Belgian Waloon Region, contract no. 1510501.

\bibliographystyle{plainnat}
\bibliography{sample}

\appendix
\section{ResNets} \label{appendixA}

\subsection{Preprocessing}
ResNet v1 has been trained on ImageNet with a vgg preprocessing. It consists of a random crop and a random flip. ResNet v2 applies the inception preprocessing that uses, on top of the vgg preprocessing, random color distortion (hue, contrast, brightness and saturation). Input image size for ResNet v1 and v2 are respectively of 224x224x3 and 299x299x3.

	\begin{figure}[H]
		\centering
		\includegraphics[scale=0.35]{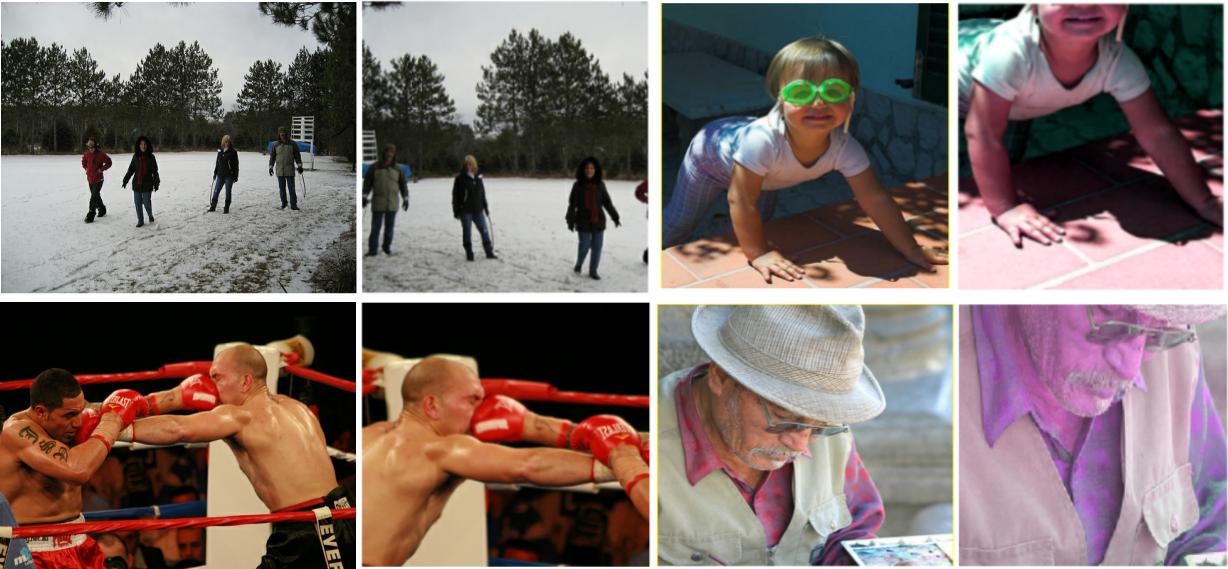}
		\caption{Top-left and bottom-left are the VGG processing. Top right and bottom right are the Inception processing}
	\end{figure}

\subsection{Parameters}
\newpage
\begin{table}[!ht]
  \caption{ResNets Parameters}
  \label{ok}
  \centering
          \def\arraystretch{1.3}%  1 is the default, change whatever you need
  \begin{tabular}{ll}
    \toprule
    Parameter     & Value      \\
    \midrule
    ResNet v1 version & ResNet-50       \\
    ResNet v2 version     & ResNet-50     \\
    ResNet v1 input size     & 224x224x3     \\
    ResNet v2 input size     & 299x299x3     \\
    ResNet v1 CBN inference     & moving average     \\
    ResNet v2 CBN inference     &  exponential moving average    \\
    CBN decay & 0.99 \\
    CBN damping factor $\epsilon$ & 1e-5 \\
    CBN MLP hidden units  & 512      \\
    
    Blocks with CBN  & all (by default)      \\

    \bottomrule
  \end{tabular}
\end{table}

\section{Sequence to sequence model}

To conduct our experiments, we use the TensorFlow \citep{abadi2016tensorflow} library as well as the google seq2seq framework \citep{Britz:2017}. We release our code on github \footnote{\url{https://github.com/jbdel/mmt_cbn}}.
We normalize and tokenize English and German descriptions using the Moses tokenizer scripts \cite{Koehn:2007}. We use the byte pair encoding algorithm on the train set to convert space-separated tokens into subwords \citep{sennrich2016subword} with 10K merge operation, reducing our vocabulary size to 5234 and 7052 words for English and German respectively. Embeddings are learned along with the model.
\subsection{Encoder}

Both encoders are equipped with layer normalization \citep{ba2016layer} where each hidden unit adaptively normalizes its incoming activations with a learnable gain and bias.

\subsection{Decoder}

We initialize the decoder hidden state $\vh_0$ of the CGRU with a non-linear transformation of the average source annotation:
\begin{equation}
\vh_0 = \tanh \bigg( \mW_{init} \cdot \frac{1}{M} \sum\limits_{i=1}^M \vh_i \bigg),  \vh_i \in h
\label{meaneq}
\end{equation}

The decoder is a conditional GRU \footnote{\url{https://github.com/nyu-dl/dl4mt-tutorial/blob/master/docs/cgru.pdf}} that consists of two stacked GRU activations called $\text{REC}_1$ and $\text{REC}_2$ and an attention mechanism $f_{\text{att}}$ in between (called ATT in the footnote paper).	At each time-step $t$, REC1 firstly computes a hidden state proposal $\vs_t$ based on the previous hidden state $\widehat{\vs}_{t-1}$ and the previous emitted word $y_{t-1}$:
	\begin{align}
	\vz_t =& ~ \sigma \left(  \mW_z \mE_Y[y_{t-1}] + \mU_z \widehat{\vs}_{t-1}  \right) \nonumber \\
	\vr_t =& ~ \sigma \left(  \mW_r \mE_Y[y_{t-1}] + \mU_r \widehat{\vs}_{t-1}  \right)  \nonumber \\ 
	\underline{\vs}_t =& ~\text{tanh} \left(   \mW \mE_Y[y_{t-1}] + \vr_t \odot (\mU\widehat{\vs}_{t-1})  \right)  \nonumber \\        
	\vs_t =& (1 - \vz_t) \odot \underline{\vs}_t + \vz_t \odot \widehat{\vs}_{t-1}
	\end{align}
	Then, the attention mechanism computes $\vc_t$ over the source sentence using the annotations sequence $h$ and the intermediate hidden state proposal $\vs_t$ (cfr. section \ref{enc_dec}).
    
   	Finally, the second recurrent cell $\text{REC}_2$, computes the hidden state $\widehat{\vs}_t$ of the $\text{cGRU}$ by looking at the intermediate representation $\vs_t$ and context vector $\vc_t$:
	\begin{align}
	\vz_t =& \sigma \left( \mW_z \vc_t + \mU_z \vs_t\right) \nonumber \\
	\vr_t =& \sigma \left( \mW_r \vc_t + \mU_r \vs_t\right) \nonumber \\
	\widehat{\underline{\vs}}_t =& \text{tanh} \left(  \mW \vc_t  + \vr_t \odot (\mU \vs_t )  \right)  \nonumber \\       
	\widehat{\vs}_t =& (1 - \vz_t) \odot \underline{\vs}_t + \vz_t \odot \vs_t
	\end{align}	
\\
\subsection{Parameters}

\begin{table}[!ht]
  \caption{Sequence to sequence parameters}
  \label{ok2}
  \centering
          \def\arraystretch{1.3}%  1 is the default, change whatever you need
  \begin{tabular}{ll}
    \toprule
    Parameter     & Value      \\
    \midrule
    Source and target embeddings & 128       \\
    GRU and CGRU Layer size    & 256     \\
    Attention size    & 256     \\
    GRU input dropout & 0.7 \\
    GRU output dropout & 0.5 \\
    CGRU input dropout & 1.0 \\
    CGRU output dropout & 1.0 \\
    Softmax output dropout \ref{eqn:9} & 0.5 \\
    Optimizer & Adam \\
    Learning rate & 0.0004 \\
    Optimize epsilon & 0.0000008 \\
    Batch-size & 32 \\
    Inference Beam-Size & 12\\
   \bottomrule
  \end{tabular}
\end{table}

\section{Further Results}

\begin{table*}[!ht]
		\centering
        		\caption{Use of CBN in the different ResNet stage (Stage 4 only still to compute for camera ready version)}
        \hspace*{-0.9cm}
        \def\arraystretch{1.3}%  1 is the default, change whatever you need
		\begin{tabular}{lcccccc}
        \multicolumn{1}{c}{\bf RN v1 CBN Pool5}  &\multicolumn{2}{c}{\bf Test 2016}  &\multicolumn{2}{c}{\bf Test 2017}  &\multicolumn{2}{c}{\bf Ambiguous COCO}
			\\ \hline \\
			&BLEU$\uparrow$&METEOR$\uparrow$ &BLEU$\uparrow$&METEOR$\uparrow$ &BLEU$\uparrow$&METEOR$\uparrow$  \\
			 \\
			All   
            &39.4$\pm$ 0.8& 57.9 $\pm$ 0.6 & 31.5 $\pm$ 0.4 & 52.2 $\pm$ 0.5 & 27.4 $\pm$ 0.9 & 48.1 $\pm$ 0.6  \\
			Stages 2 - 4
            &38.7 $\pm$ 0.6& 57.0 $\pm$ 0.7 & 31.4$\pm$ 0.9 & 51.4 $\pm$ 0.4 & 27.3 $\pm$ 0.9 & 46.8 $\pm$ 0.9  \\
			Stages 3 - 4
            &38.8 $\pm$ 0.8& 56.1 $\pm$ 0.7 & 30.4 $\pm$ 0.8 & 51.1 $\pm$ 0.6 & 25.9 $\pm$ 1.0 & 46.0 $\pm$ 0.6  \\
			\hline \\

		\end{tabular}      
        \label{score-tabular2}
	\end{table*}%

\end{document}